\documentclass[conference]{IEEEtran}
\IEEEoverridecommandlockouts
\usepackage{cite}
\usepackage{amsmath,amssymb,amsfonts}
\usepackage{algorithmic}
\usepackage{graphicx}
\usepackage{textcomp}
\usepackage{xcolor}
\usepackage{caption}
\usepackage{subfigure}
\usepackage{multicol}
\usepackage{multirow}
\usepackage{float}
\def\BibTeX{{\rm B\kern-.05em{\sc i\kern-.025em b}\kern-.08em
    T\kern-.1667em\lower.7ex\hbox{E}\kern-.125emX}}
\begin{document}
\bstctlcite{IEEEexample:BSTcontrol}

\title{Two-stream Multi-dimensional Convolutional Network for Real-time Violence Detection}

\author{\IEEEauthorblockN{Dipon Kumar Ghosh and Amitabha Chakrabarty}
\IEEEauthorblockA{\textit{Department of Computer Science and Engineering, BRAC University, Dhaka, Bangladesh}}
}

\maketitle
\thispagestyle{plain}
\pagestyle{plain}





\begin{abstract}
The increasing number of surveillance cameras and security concerns have made automatic violent activity detection from surveillance footage an active area for research. 
Modern deep learning methods have achieved good accuracy in violence detection and proved to be successful because of their applicability in intelligent surveillance systems.
However, the models are computationally expensive and large in size because of their inefficient methods for feature extraction.
This work presents a novel architecture for violence detection called Two-stream Multi-dimensional Convolutional Network (2s-MDCN), which uses RGB frames and optical flow to detect violence. 
Our proposed method extracts temporal and spatial information independently by 1D, 2D, and 3D convolutions. 
Despite combining multi-dimensional convolutional networks, our models are lightweight and efficient due to reduced channel capacity, yet they learn to extract meaningful spatial and temporal information.
Additionally, combining RGB frames and optical flow yields 2.2\% more accuracy than a single RGB stream.
Regardless of having less complexity, our models obtained state-of-the-art accuracy of 89.7$\%$ on the largest violence detection benchmark dataset. 

\end{abstract}


\section{Introduction}	
    \label{sec:introduction}

There are several advantages to detecting violent activity from surveillance video.
In today's world, security cameras may be found in practically every public area such as an office, hospitals, educational institutes, and shopping malls, among other places. As the number of security cameras grows, so does the demand for more sophisticated methods of monitoring the footages. Using manual methods to monitor and identify violence in real-time from video footage is time-consuming and costly. A further disadvantage is that it may take some time to notify the appropriate authorities responsible for taking action in the event of an emergency, while an automated violence detection system may do so nearly instantaneously.

When it comes to violence detection, it may be regarded as a subset of human action recognition, which seeks to recognize conventional human activity \cite{cheng2019rwf}.
In image recognition, $I(h,w)$, the spatial characteristics $h$ and $w$ that offer information about the scene are generally extracted. In contrast, there is another dimension in video data $I(t,h,w)$ that contains information about the passage of time, which reveals the changes in spatial features with time.
It is required to extract both spatial and temporal information to detect violence from video data. 

\begin{figure}[t]
    \centering
    \subfigure[Movies fight detection dataset samples.]
    {
        \includegraphics[width=0.4\textwidth]{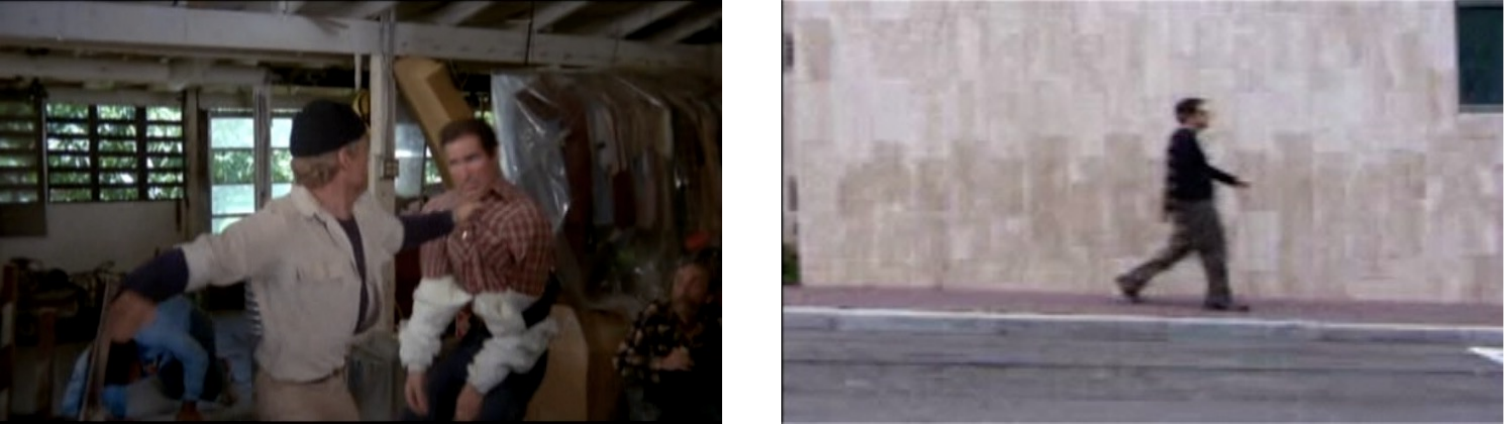}
        \label{fig:third_sub}
    }
    
    \subfigure[Hockey fight detection dataset samples.]
    {
        \includegraphics[width=0.4\textwidth]{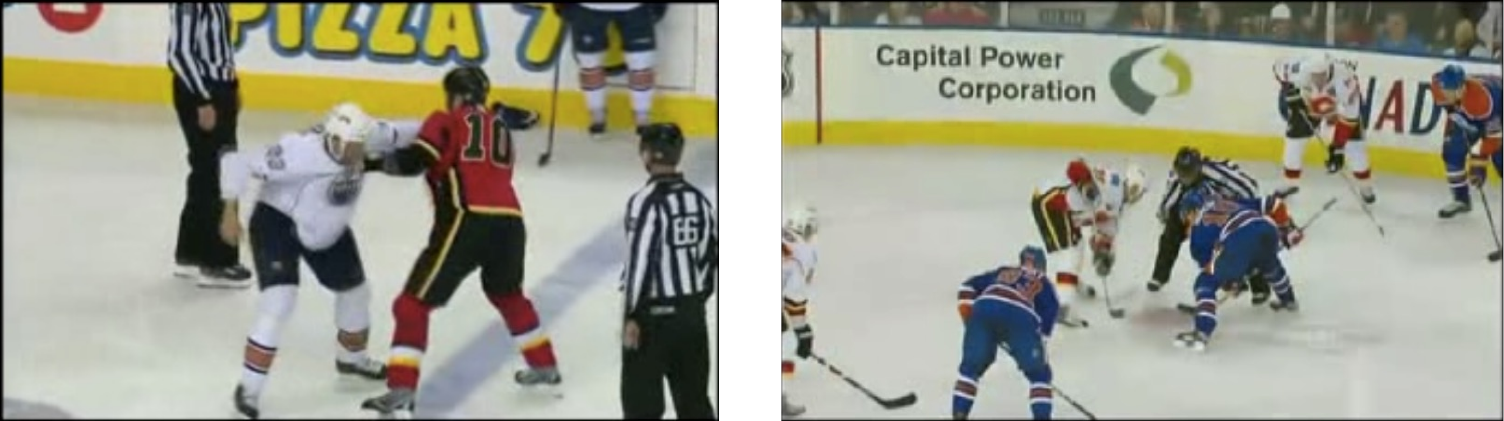}
        \label{fig:second_sub}
    }
    
    \subfigure[RWF-2000 violence detection dataset samples.]
    {
        \includegraphics[width=0.4\textwidth]{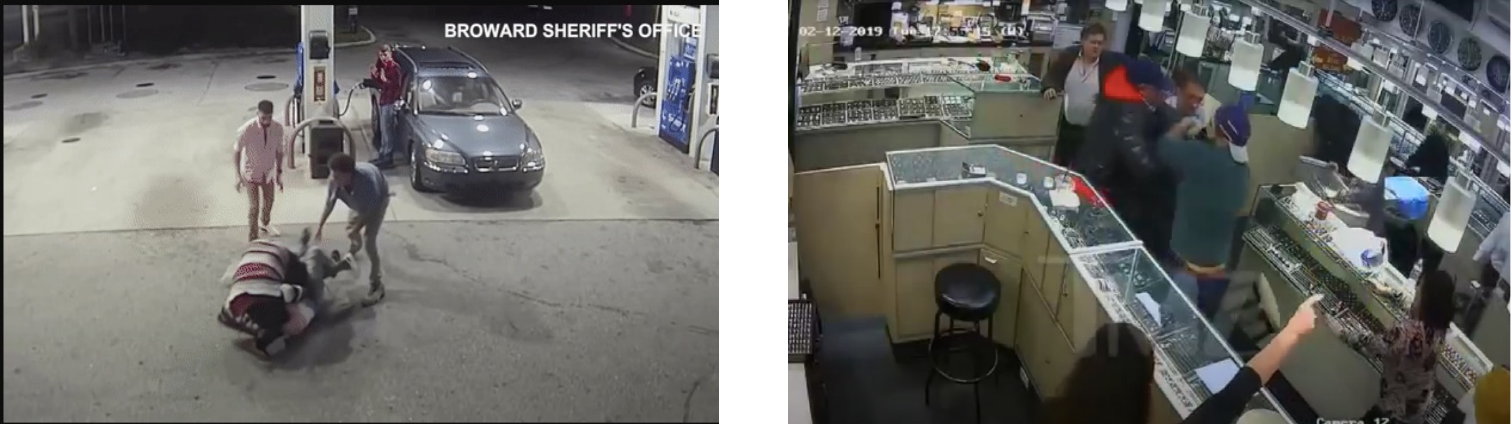}
        \label{fig:first_sub}
    }
    \caption{Samples from different violent detection dataset.}
    \label{fig:sample_dataset}
\end{figure}

Previously, researchers employed a variety of feature extraction approaches, such as ViF~\cite{hassner2012violent_6}, STIPs~\cite{de2010violence_7}, iDT~\cite{wang2013action_9}, and fed the results to classic classification models, such as support vector machine (SVM).
Real-world settings, on the other hand, are complicated, as seen in Fig.~\ref{fig:sample_dataset}, and it is difficult to extract relevant information from hand-crafted feature descriptors.
Convolutional neural network (CNN), gated  recurrent  unit (GRU),  and  long  short-term  memory  (LSTM)  network  are deep learning-based approaches that have recently been demonstrated to be effective in learning interpretable and robust features from images, identifying spatial information, and achieving cutting edge results on image classification, segmentation, and other computer vision tasks~\cite{tan2019efficientnet,lin2017fpn}. 
Research has been carried out to extend the success of deep learning approaches to video analysis, with results that are at the state-of-the-art the field~\cite{eco14zolfaghari2018, 3dres101_hara2018can}.

The significance of extracted spatial and temporal features determines the success in violence detection as well as general human action recognition.
Modern approaches, in addition to RGB frames, depend on optical flow for temporal information. 
Furthermore, because of the high cost of computational complexity and the vast number of parameters, they are inefficient for any real-world application.
We believe that the problem lies in the way spatial and temporal characteristics are extracted in a current deep-learning model.
When developing the models, generally spatial and temporal characteristics are retrieved in a sequential manner, with temporal features extraction occurring after spatial data extraction. This results in a loss of information due to the fact that the features are not extracted from the same spatio-temporal position throughout the procedure.




	
	\begin{figure*}[!htb]
	\centering
	\includegraphics[width=0.9\linewidth]{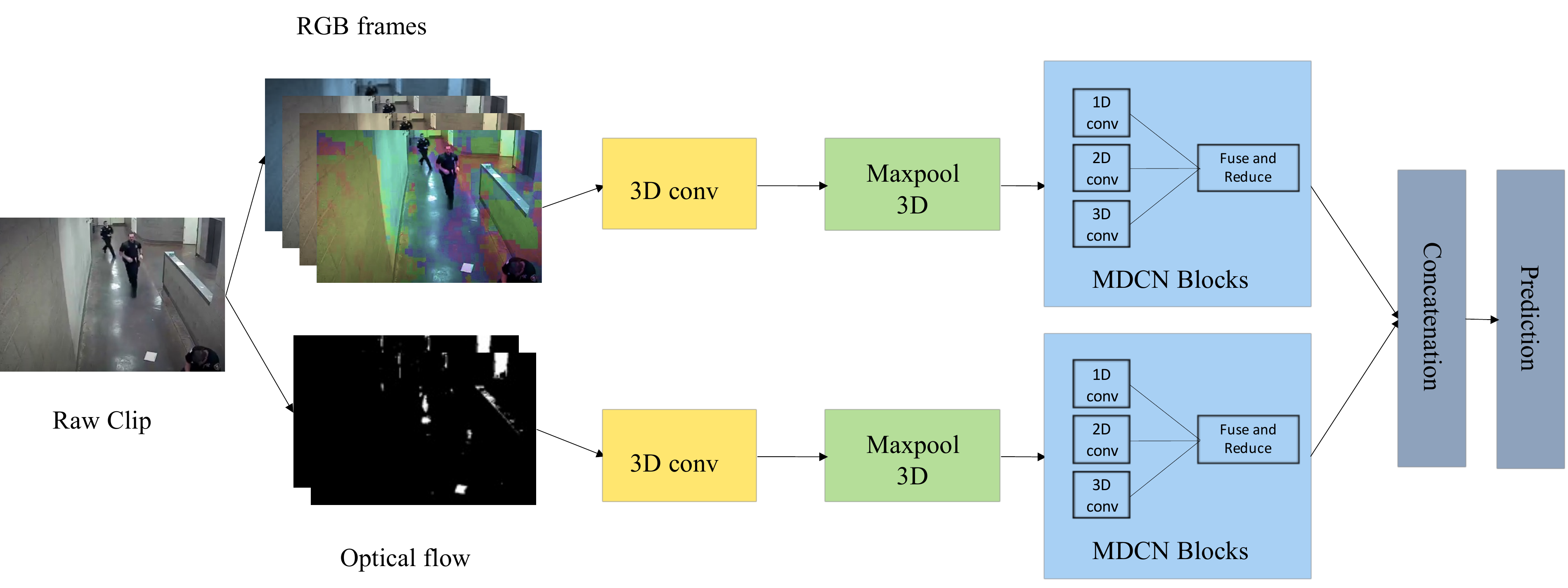}
	\caption{Overall pipeline of 2s-MDCN.}
	\label{fig:pipeline}
	\end{figure*}    

To address this problem, we propose a novel architecture for violence detection named the two-stream multi-dimensional convolutional network (2s-MDCN), as illustrated in Fig. ~\ref{fig:pipeline}.
Our proposed 2s-MDCN uses both RGB and optical flow as input and thus, the model has two separate branches for RGB stream and optical flow stream.
Our proposed model employs a fusion of multi-dimensional convolution processes to filter out spatial and temporal data separately retrieved from the same spatio-temporal point, therefore minimizing information loss.
Our proposed method employs optical flow along with the RGB frames.
To demonstrate the effectiveness of our model, we conduct extensive experiments on three benchmark datasets for violence detection, including RWF-2000 violence ~\cite{cheng2019rwf}, Hockey-fight~\cite{hockey_nievas2011violence} and Movies-fight~\cite{movie_nievas2011violence} dataset.
We also performed experiments with a single RGB stream and optical flow stream. Our experiments show that adding optical flow increased the model accuracy by a significant amount.
Despite having low parameters and less complexity, our model obtained state-of-the-art accuracy in violence detection benchmark datasets.

The primary contributions of this work can be summarized as follows:
\begin{itemize}
    \item We explored how important it is to coordinate temporal and spatial features in violence detection.
    \item We proposed 2s-MDCN, a novel architecture for violence detection which takes RGB frames and optical flow as input. Our model extracts temporal and spatial features independently of each other whereas, current models do that in a sequential fashion which may not be suitable to fully exploit the features.
    \item Finally, this work can be considered a strong baseline for violence detection as our model achieved state-of-the-art accuracy in the largest violence detection dataset. We hope our baseline is helpful for future work in violent action detection.
\end{itemize}	
\section{Related work}
    \label{sec:relatedwork}
    \subsection{Traditional Methods}
    
    Violence can be detected by traditional methods using hand-crafted feature extraction algorithms and using classical machine learning algorithms such as KNN, AVM, and Adaboost as a classifier. 
    As done by, Harris corner detector~\cite{chen2008recognition}, improved dense trajectory (iDT)~\cite{wang2013action_9}, motion scale-invariant feature transform (MoSIFT)~\cite{xu2014violent_22}, space-time interest points (STIP)~\cite{de2010violence_7}.
    
    Hassner et al.~\cite{hassner2012violent_6} used optical flow magnitude series to detect violence in videos. The features are called violent flow (ViF) descriptors.
    Later, their method was improved by introducing the orientation in the violent flow (ViF) descriptors~\cite{gao2016violence_23}.
    Improved dense trajectory features iDT~\cite{wang2013action_9} remarkably improved the accuracy of human action recognition. 
    This method along with the fisher encoding method extracts significant spatio-temporal features from violent videos~\cite{bilinski2016human_24}.
    
    \subsection{Deep Learning Based Methods}
    Recently, deep learning-based methods gained much interest due to their improved accuracy and better performance than traditional methods. 
    Convolutional neural network (CNN)~\cite{lecun1995convolutional}, 3D convolutional network~\cite{ji_3dconv},  and long short term memory (LSTM) network~\cite{10.1162/neco.1997.9.8.1735} are widely used architecture for the purpose of video understanding~\cite{11convlstm_donahue2015long, c3dtran2015learning, i3dcarreira2017quo}.
    Different deep learning-based approaches have also been used in violence detection~\cite{26_zhou2017violent, 29_sudhakaran2017learning}. 
    	
	Three input streams, RGB frames, optical flow, and accelerated flow maps, were used as input for violence detection~\cite{28_dong2016multi}.
	They used LSTM network to filter out features from the input streams. 
	Using their individual benefits, CNN and LSTM are used together in violent detection~\cite{30_shi2015convolutional,29_sudhakaran2017learning}. 
	The spatial features are extracted by the convolution layers and the mapping of temporal frames was done by LSTM layers.
	FightNet~\cite{26_zhou2017violent} used features from multiple streams, such as RGB frames, optical flow, and acceleration images, and fused them for violence detection. In~\cite{cheng2019rwf}, Flow-Gate network was proposed which used fusion of RGB frames and optical flow.	
\section{The Proposed 2s-MDCN}
    \label{mdcn}
    
    \begin{figure}[!htb]
	\centering
	\includegraphics[width=0.9\linewidth]{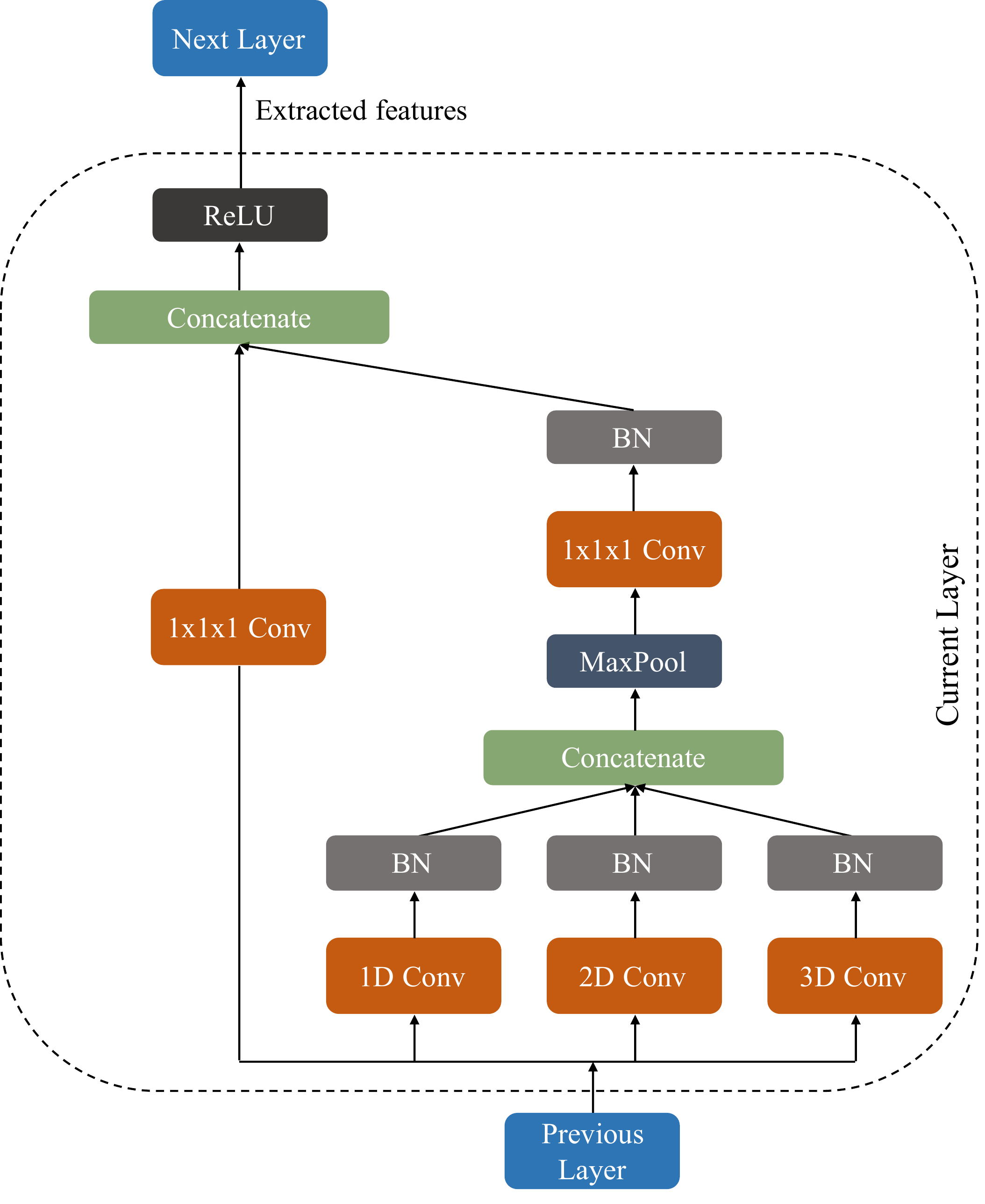}
	\caption{A single multi-dimensional convolutional block.}
	\label{fig:mdcn_block}
	\end{figure}
    
    In this section, we discuss the architecture of our model in detail. 
    Figure~\ref{fig:pipeline} shows the overall pipeline of our model. 
    Raw RGB frames captured from cameras and optical flow are used as input to the model.
    First, the input is passed through a 3D convolutional layer which performs a 3D convolution operation and reduces the size of spatial dimension to prepare for the multi-dimensional convolutional (MDC) blocks.
    Two separate branches are used for RGB frames and optical flow, which are concatenated after the feature extraction from both branches.
    Each block in the model extracts spatial and temporal features independently from each other and detects violence.
    


    
    \subsection{Multi-dimensional Convolutional (MDC) Block}
	The input to each one of the blocks is of shape $ {C}\times{D}\times{H}\times{W} $, where ${C}$ is the number of channels, ${D}$ represents the frame numbers, ${H}$ and ${W}$ denotes height and width respectively. 
	The values in $ {H}\times{W} $ contain spatial information of a particular frame while $ {D} $ frames contain temporal information for the corresponding pixel.
	As figure ~\ref{fig:mdcn_block} illustrates, each block of our proposed model contains three main convolutional layers for extracting spatial and temporal features.
    
    \textbf{ 1D convolutional layer:} The first module is 1D convolutional layer which extracts temporal features from the input.
    The kernel size of this convolutional layer is $ {{k}_{t}}\times{1}\times{1} $. 
    So that it performs convolution operation on a particular pixel over $ {{k}_{t}} $ frames and extracts only temporal information of the corresponding pixel.
    
    \textbf{ 2D convolutional layer:} The second module is 2D convolutional layer which extracts temporal features from the input.
    The kernel size of this convolutional layer is $ {1}\times{{k}_{s}}\times{{k}_{s}} $. 
    So that it performs convolution operation on a single frame and extracts only spatial information from the particular frame.
    
   \textbf{ 3D convolutional layer:} The second module is 3D convolutional layer which extracts both spatial and temporal features from the input.
    The kernel size of this convolutional layer is $ {{k}_{t}}\times{{k}_{s}}\times{{k}_{s}} $. 
    So that it performs convolution operation over ${{k}_{t}}$ frames and extracts temporal and spatial frames together.
    
    Each of these three modules is followed by a batch normalization layer.
    In our model, we used $ 3 $ the value of $ {{k}_{t}} $ and $ {{k}_{s}} $.
    They extract information independently from the same spatio-teporal position of  the input and fuse together.
    In this process, our model can extract all the information from the input which reduces information loss and improves accuracy.
    Then, the 
	fused features are passed through a 
    $maxpooling$ layer and a $ {1}\times{1}\times{1} $ convolution layer which reduces the number of channels for the next layer.
    This reduction module keeps our model size small.
    Moreover, a concatenated skip connection~\cite{huang2017densely}, followed by a ReLU layer, is added to stabilize the model which also helped to improve accuracy.
    Skip connection concatenates features from the previous layers to the current layer which allows more information to be obtained from the previous layers and reduces loss of information. Concatenated skip connections also help the gradient to propagate better and fix the vanishing gradient problem.
    To match the input channel to the output channel, we have used a convolutional layer of kernel size ${1}\times{1}\times{1}$ and stride of size ${1}\times{2}\times{2}$.




	\begin{figure}[!htb]
	\centering
	\includegraphics[width=0.9\linewidth]{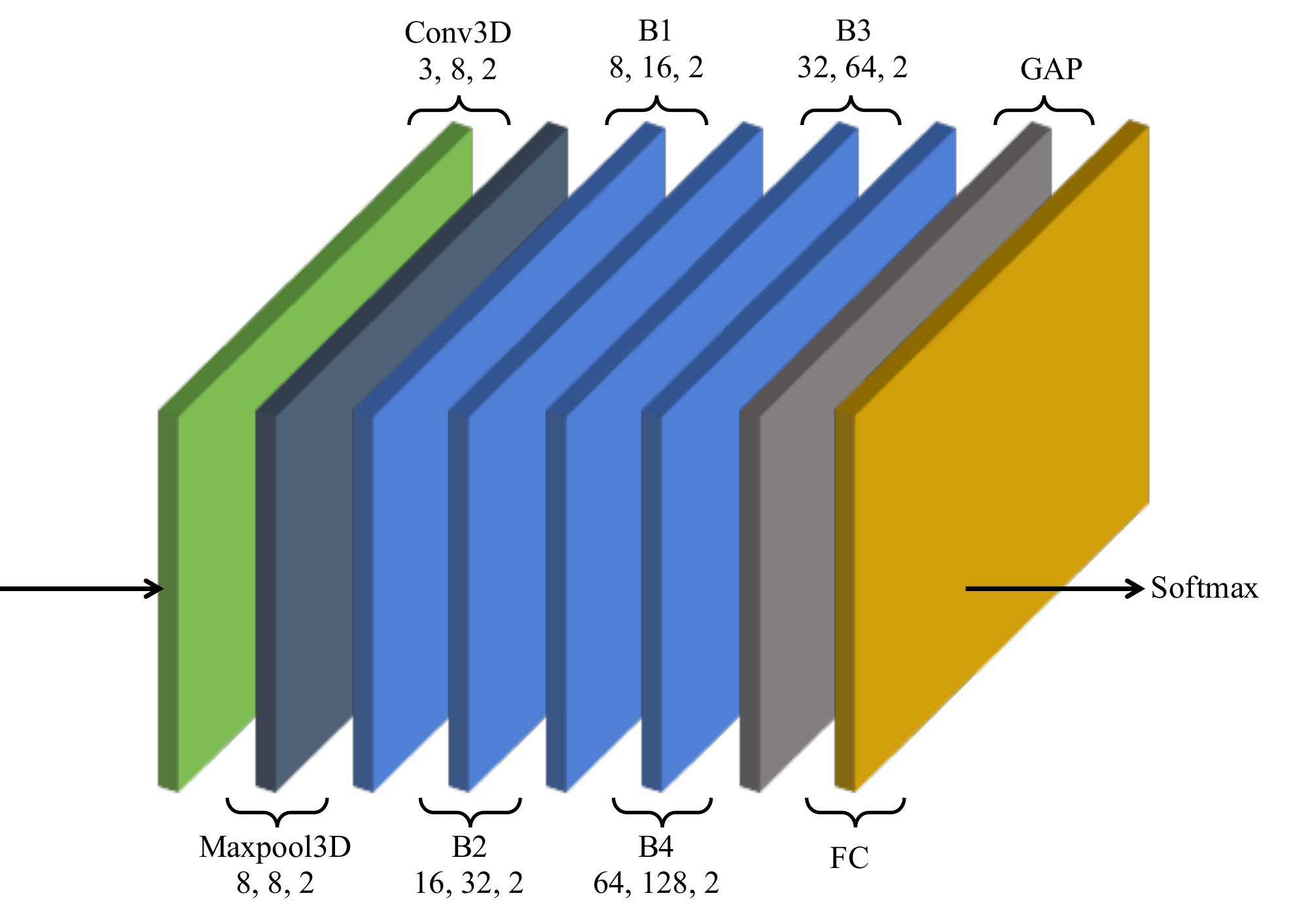}
	\caption{Layers of 2s-MDCN.}
	\label{fig:full_network}
	\end{figure}      
    
\begin{table*}[]
\centering
\begin{tabular}{|cc|cc|cc|}
\hline
\multicolumn{2}{|c|}{\multirow{2}{*}{Layers}}                                                   & \multicolumn{2}{c|}{2s-MDCN}               & \multicolumn{2}{c|}{\multirow{2}{*}{Output Size}} \\ \cline{3-4}
\multicolumn{2}{|c|}{}                                                                          & \multicolumn{1}{c|}{RGB} & optical flow & \multicolumn{2}{c|}{}                             \\ \hline
\multicolumn{2}{|c|}{\multirow{2}{*}{input}}                                                    & \multicolumn{2}{c|}{\multirow{2}{*}{}}  & \multicolumn{2}{c|}{$3\times32\times{224}^2$(RGB)}       \\ 
\multicolumn{2}{|c|}{}  & \multicolumn{2}{c|}{} & \multicolumn{2}{c|}{$2\times32\times{224}^2$(optical flow)}               \\ \hline
\multicolumn{2}{|c|}{conv$_1$}  & \multicolumn{2}{c|}{conv 3D: ${5}\times{{7}^2} $, stride:  ${1}\times{{2}^2} $}               & \multicolumn{2}{c|}{$8\times{32}\times{112}^2$} \\ \hline

\multicolumn{2}{|c|}{pool$_1$}  & \multicolumn{2}{c|}{max pool: ${1}\times{{3}^2}$, stride:  ${1}\times{{2}^2} $}               & \multicolumn{2}{c|}{$8\times{32}\times{56}^2$} \\ \hline

\multicolumn{1}{|c|}{\multirow{4}{*}{mdcn$_1$}} & \multicolumn{1}{c|}{\multirow{3}{*}{conv layer}} & \multicolumn{2}{c|}{conv 1D: ${3}\times{{1}^2}$}   & \multicolumn{2}{c|}{\multirow{4}{*}{$16\times{32}\times{28}^2$}}      \\
\multicolumn{1}{|c|}{}                       & \multicolumn{1}{c|}{}                            & \multicolumn{2}{c|}{conv 2D: ${1}\times{{3}^2} $}           & \multicolumn{2}{c|}{}                             \\
\multicolumn{1}{|c|}{}                       & \multicolumn{1}{c|}{}                            & \multicolumn{2}{c|}{conv 3D: ${3}\times{{3}^2} $}           & \multicolumn{2}{c|}{}                             \\ \cline{2-4}

\multicolumn{1}{|c|}{}                                     & \multirow{2}{*}{fuse and reduce} & \multicolumn{2}{c|}{max pool: ${1}\times{{3}^2} $, stride:  ${1}\times{{2}^2} $ }            & \multicolumn{2}{c|}{}     \\
\multicolumn{1}{|c|}{}   &       & \multicolumn{2}{c|}{conv: ${1}\times{{1}^2} $ }  & \multicolumn{2}{c|}{} \\ \hline

\multicolumn{1}{|c|}{\multirow{4}{*}{mdcn$_2$}} & \multicolumn{1}{c|}{\multirow{3}{*}{conv layer}} & \multicolumn{2}{c|}{conv 1D: ${3}\times{{1}^2}$}   & \multicolumn{2}{c|}{\multirow{4}{*}{$32\times{32}\times{28}^2$}}      \\
\multicolumn{1}{|c|}{}                       & \multicolumn{1}{c|}{}                            & \multicolumn{2}{c|}{conv 2D: ${1}\times{{3}^2} $}           & \multicolumn{2}{c|}{}                             \\
\multicolumn{1}{|c|}{}                       & \multicolumn{1}{c|}{}                            & \multicolumn{2}{c|}{conv 3D: ${3}\times{{3}^2} $}           & \multicolumn{2}{c|}{}                             \\ \cline{2-4}

\multicolumn{1}{|c|}{}                                     & \multirow{2}{*}{fuse and reduce} & \multicolumn{2}{c|}{max pool: ${1}\times{{3}^2} $, stride:  ${1}\times{{2}^2} $ }            & \multicolumn{2}{c|}{}     \\ 
\multicolumn{1}{|c|}{}   &       & \multicolumn{2}{c|}{conv: ${1}\times{{1}^2} $ }  & \multicolumn{2}{c|}{} \\ \hline

\multicolumn{1}{|c|}{\multirow{4}{*}{mdcn$_3$}} & \multicolumn{1}{c|}{\multirow{3}{*}{conv layer}} & \multicolumn{2}{c|}{conv 1D: ${3}\times{{1}^2}$}   & \multicolumn{2}{c|}{\multirow{4}{*}{$64\times{32}\times{14}^2$}}      \\
\multicolumn{1}{|c|}{}                       & \multicolumn{1}{c|}{}                            & \multicolumn{2}{c|}{conv 2D: ${1}\times{{3}^2} $}           & \multicolumn{2}{c|}{}                             \\
\multicolumn{1}{|c|}{}                       & \multicolumn{1}{c|}{}                            & \multicolumn{2}{c|}{conv 3D: ${3}\times{{3}^2} $}           & \multicolumn{2}{c|}{}                             \\ \cline{2-4}

\multicolumn{1}{|c|}{}                                     & \multirow{2}{*}{fuse and reduce} & \multicolumn{2}{c|}{max pool: ${1}\times{{3}^2} $, stride:  ${1}\times{{2}^2} $ }            & \multicolumn{2}{c|}{}     \\ 
\multicolumn{1}{|c|}{}   &       & \multicolumn{2}{c|}{conv: ${1}\times{{1}^2} $ }  & \multicolumn{2}{c|}{} \\ \hline

\multicolumn{1}{|c|}{\multirow{4}{*}{mdcn$_4$}} & \multicolumn{1}{c|}{\multirow{3}{*}{conv layer}} & \multicolumn{2}{c|}{conv 1D: ${3}\times{{1}^2}$}   & \multicolumn{2}{c|}{\multirow{4}{*}{$128\times{32}\times{7}^2$}}      \\
\multicolumn{1}{|c|}{}                       & \multicolumn{1}{c|}{}                            & \multicolumn{2}{c|}{conv 2D: ${1}\times{{3}^2} $}           & \multicolumn{2}{c|}{}                             \\
\multicolumn{1}{|c|}{}                       & \multicolumn{1}{c|}{}                            & \multicolumn{2}{c|}{conv 3D: ${3}\times{{3}^2} $}           & \multicolumn{2}{c|}{}                             \\ \cline{2-4}

\multicolumn{1}{|c|}{}                                     & \multirow{2}{*}{fuse and reduce} & \multicolumn{2}{c|}{max pool: ${1}\times{{3}^2} $, stride:  ${1}\times{{2}^2} $ }            & \multicolumn{2}{c|}{}     \\ 
\multicolumn{1}{|c|}{}   &       & \multicolumn{2}{c|}{conv: ${1}\times{{1}^2} $ }  & \multicolumn{2}{c|}{} \\ \hline

\multicolumn{4}{|c|}{concatenate, global average pool, fc}                                                                                & \multicolumn{2}{c|}{\#classes (2)}                \\ \hline
\end{tabular}
\caption{ Overall architecture of 2s-MDCN.}

    \label{tab:network_table}
\end{table*}

    \subsection{Multi-dimensional Convolutional Network}
   As figure ~\ref{fig:full_network} illustrates, our proposed network consists of $ 4 $ MDC blocks and a 3D convolutional layer at the beginning. 
   Both RGB and optical flow streams have identical numbers of MDC blocks.
   The first 3D convolutional layer is of kernel size $ {5}\times{7}\times{7} $ and stride of size $ {1}\times{2}\times{2} $.
   This layer is followed by a $ maxpool $ layer with stride value of $ {1}\times{2}\times{2} $.
   The convolution layer and $maxpool$ layer are used to reduce the size of spatial dimension.
   The output from these layers is passed into the $ 4 $ MDC blocks, where temporal and spatial information are extracted.
   The numbers of channels of output from the convolution blocks are $8$, $16$, $32$, $64$, and $128$.
   After extracting features from RGB and optical flow stream, they are concatenated and a $global$ $average$ $pooling$ layer is used to combine and reduce the extracted feature and finally, a fully connected (FC) and $softmax$ layer is used to detect violence.
   Table~\ref{tab:network_table} displays the kernel size, stride size, and number output channel of each block in detail.
   The shape of convolutional kernel is denoted by $ {T}\times{S}^2 $, where $T$ kernel temporal dimension and $S$ indicated spatial dimension. Stride is also represented in the same manner (temporal stride, spatial stride$^2$). The output size is represented by channel$\times$temporal length$\times$spatial dimension$^2$. The features are passed into $conv$ $layers$ independently and then concatenated and passed into $fuse$ $and$ $reduce$ layer.

\section{Experiments}
    In this section, we analyze the performance of our model on public violence detection datasets. Our model outperforms the previous models and achieves state-of-the-art accuracy in spite of having low computational complexity and less memory footprint.

    \subsection{Datasets}
    We used three benchmark violence detection datasets for training and validating our model, namely  RWF-200 violence dataset~\cite{cheng2019rwf}, Hockey-fight dataset~\cite{hockey_nievas2011violence} and Movies-Fight dataset~\cite{movie_nievas2011violence}. 
    At the moment, the RWF-2000 violence dataset is the largest dataset for violence detection with 2000 clips. 1600 clips out are kept for training and 400 are used for validation. The clips from the dataset are collected from real-world surveillance cameras. The number of characters in the clips are not fixed, dynamic characteristics vary a lot and the background is complicated.
    Hockey-fight dataset is collected from video footage of hockey games. There are 1000 clips in the dataset, where half of them are violent and the rest are non-violent. Movies-fight dataset consists of clips from movies with a total of 200 clips. 
    
    \begin{figure*}[]
	\centering
	\includegraphics[width=0.9\linewidth]{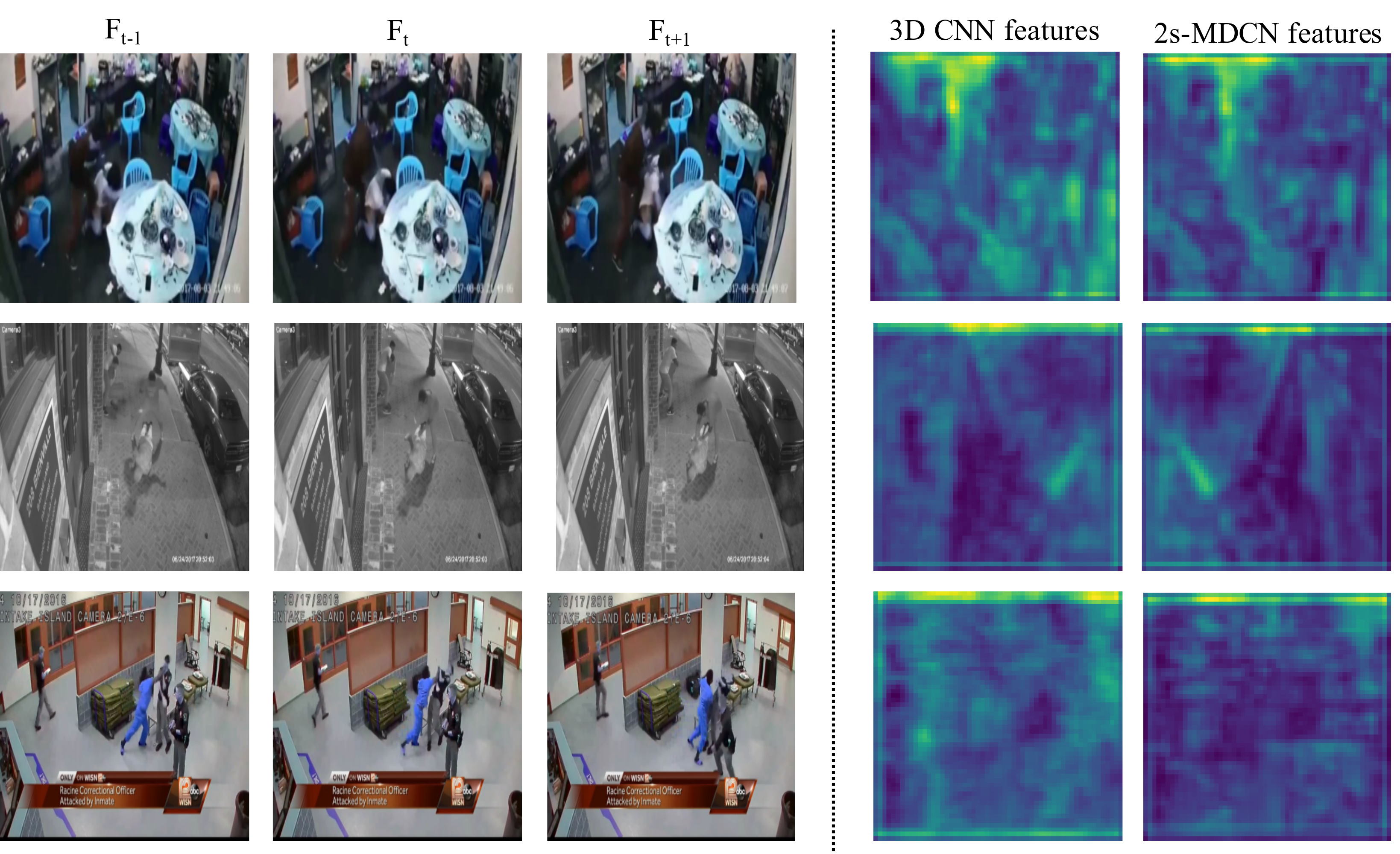}
	\caption{Visualization of extracted features from 2s-MDCN.}
	\label{fig:features_vis}
	\end{figure*}

    \subsection{Training details}
     For training, we sampled $32$ frames from each clip and each frame was resized to ${224}\times{224}$. 
     Thus, our input size for the was ${3}\times{32}\times{224}\times{224}$. 
     Following the procedure of~\cite{cheng2019rwf}, we used brightness transformation and random rotation to augment our data in order to prevent overfitting.
     We implemented our model by using PyTorch~\cite{pytorchpaszke2019} which is a deep-learning framework. Stochastic gradient descent (SGD) optimizer was used to train our network with nesterov momentum~\cite{nesterovbotev2017} of value $0.9$. The value of weight decay was set to ${1e}^{-3}$. The initial learning rate was set to $0.1$, which was reduced by a factor of $10$ after $25$ and $75$ epochs. We trained our model for $100$ epochs with a batch size of $16$. 
     Our model was trained from scratch with the corresponding datasets.
     



\section{Results}
    In this section, we report the outcome of the experiments which we performed with our proposed model. This includes extensive ablation study and  comparison with current state-of-the-art models. All of the evidence points to our model being competitive with other state-of-the-art models, as well as lightweight and cost-effective.
     
     
        \begin{table}[htp]
    \begin{center}
		\begin{tabular}{lc}
			\hline
			Methods     &  Accuracy ($\%$)    \\
			\hline
            2s-MDCN(without skip connection)    &      86.75  \\
            2s-MDCN(with skip connection)    &      87.50 \\
            \hline
		\end{tabular}
    \end{center}
	\caption{Ablation study for the impact of concatenated skip connections on 2s-MDCN.}
    \label{tab:abl_skip}
	\end{table}

		\begin{table}[htp]
    \begin{center}
		\begin{tabular}{lcc}
			\hline
			Methods    &  Accuracy ($\%$)    \\
			\hline
            2s-MDCN (flow only)   &  80.50  \\
            2s-MDCN (RGB only)    &  87.50 \\
            2s-MDCN (fusion)   & 89.70 \\
            \hline
		\end{tabular}
    \end{center}
	\caption{Ablation study of the performance of 2s-MDCN with respect to input type.}
    \label{tab:abl_frame}
	\end{table}
	



	\begin{table}[htp]
    \begin{center}
		\begin{tabular}{lcc}
			\hline
			Methods     &  Frame Length  &  Accuracy ($\%$)    \\
			\hline
            2s-MDCN   &    16  &  86.30  \\
            2s-MDCN    &    32 &  89.7 \\
            \hline
		\end{tabular}
    \end{center}
	\caption{Ablation study of the performance of 2s-MDCN with respect to frame length.}
    \label{tab:abl_stream}
	\end{table}

\begin{table*}[htb]
    \begin{center}
		\begin{tabular}{lccc}
			\hline
			Methods	                     &\# Params (M)              & Complexity (GFLOPS)                & Accuracy($\%$)		\\
			\hline
            R(2+1)D~\cite{r21dtran2018closer}          					& 33.20 									& 42.42 							  						& 81.25								\\
			C3D~\cite{c3dtran2015learning} 									& 79.90 									& 38.62							   						& 82.75								\\
			ConvLSTM~\cite{29_sudhakaran2017learning} 						& -                                      	& -															& 77.00					\\
			I3D (RGB)~\cite{i3dcarreira2017quo}							& 12.30										& 111.30 													&	85.75						\\
           I3D 	(flow)~\cite{i3dcarreira2017quo}								& 12.30										& 102.52 												&	75.50						\\
           I3D (two-stream)~\cite{i3dcarreira2017quo}	 				& 24.40									& 213.85 												&	81.50						\\
           FlowGate (RGB)~\cite{cheng2019rwf}				& 0.25									& 8.76													& 84.50								\\
           FlowGate (flow)~\cite{cheng2019rwf}				& 0.25									& 8.29													&	75.50							\\
           FlowGate (fusion)~\cite{cheng2019rwf}			 	& 0.27										& 16.98													& 87.25								\\
            \hline
			\textbf{2s-MDCN (flow only)}					& \textbf{0.47}						& \textbf{4.47}								 		& \textbf{80.50}							\\
			\textbf{2s-MDCN (RGB only)}						& \textbf{0.47}						& \textbf{4.47}								 		& \textbf{87.50}							\\
			\textbf{2s-MDCN (fusion)}						& \textbf{0.94}						& \textbf{8.16}								 		& \textbf{89.70}							\\
			\hline
		\end{tabular}		
    \end{center}
	\caption{Comparison of our model with other state-of-the-art methods on RWF-2000 violence dataset.}
    \label{tab:rwf-2000}
	\end{table*}
 \begin{table}[t]
 \caption{Comparison of our model with other state-of-the-art methods on Hockey-fight and Movies-fight dataset.}
    \begin{center}
		\begin{tabular}{lcc}
			\hline
			\multirow{2}{*}{Methods}	 & \multicolumn{2}{c}{Accuracy($\%$)} \\\cline{2-3}
						     									& Hockey   				& 	Movies    \\
			\hline
            ViF~\cite{hassner2012violent_6}           										& 82.9 									& - 								\\
            LHOG+LOF~\cite{lhog_zhou2018violence} 									& 95.1 									& - 								\\
            HOF+HIK~\cite{hog_nievas2011violence} 										& 88.6									& 59.0 								\\
            HOF+HIK~\cite{hog_nievas2011violence} 										& 91.7										& 49.0 								\\
            MoWLD+BoW~\cite{mowd_zhang2017mowld} 								& 91.9										& - 								\\
            MoSFIT+HIK~\cite{hog_nievas2011violence} 								& 90.9									& 89.5							\\
            \hline
             FightNet~\cite{26_zhou2017violent}										& 97.0									& 100							\\
            3D ConvNet~\cite{3dconvnet_song2019novel} 									& 99.62									& 99.9 							\\
            ConvLSTM~\cite{29_sudhakaran2017learning} 									& 97.1										& 100 								\\
            C3D~\cite{c3dtran2015learning} 												& 96.5									& 100 								\\
			I3D (RGB)~\cite{i3dcarreira2017quo} 										& 98.5									& 100 									\\
			I3D (Flow)~\cite{i3dcarreira2017quo} 										& 84.0									& 100 									\\
			FlowGate~\cite{cheng2019rwf}										& 98.0									& 100                               \\
			\hline
			2s-MDCN									& \textbf{99.0}						& \textbf{100}									\\
			\hline
		\end{tabular}		
    \end{center}
    \label{tab:hockey_movie}
	\end{table}

    \subsection{Ablation Study}
    \label{sec:ablation}
    We performed different ablation studies in order to identify the best hyperparameters, input features and model architecture. 
    
    First, we compared our model without skip connection.  
    Table~\ref{tab:abl_skip} displays the model accuracy with and without the concatenation of skip connection. 
    Our model achieves ${89.7\%}$ with skip connection and while we turned off the skip connection it drops at ${88.7\%}$.
    The results show that skip connection helps information flow more efficiently and thus helps to improve accuracy.
    
    We also demonstrate the capability of our model with a single input stream. As shown in Table~\ref{tab:abl_stream}, our model achieves 89.7\% accuracy when two different input methods are used. We achieve 87.5\% accuracy while using only RGB stream, and by using only optical flow our model is able to achieve 78.5\% accuracy in RWF-2000 violence detection dataset.
        
    Second, as shown in Table~\ref{tab:abl_frame}, we experimented with different frame lengths and found that frame length of $32$ gives the best accuracy for our model. 
    When we used $16$ frames as input we achieved a score of ${85.3\%}$. 



\begin{figure}
    \centering
    \subfigure[Comparison between training and validation accuracy.]
    {
        \includegraphics[width=0.9\linewidth]{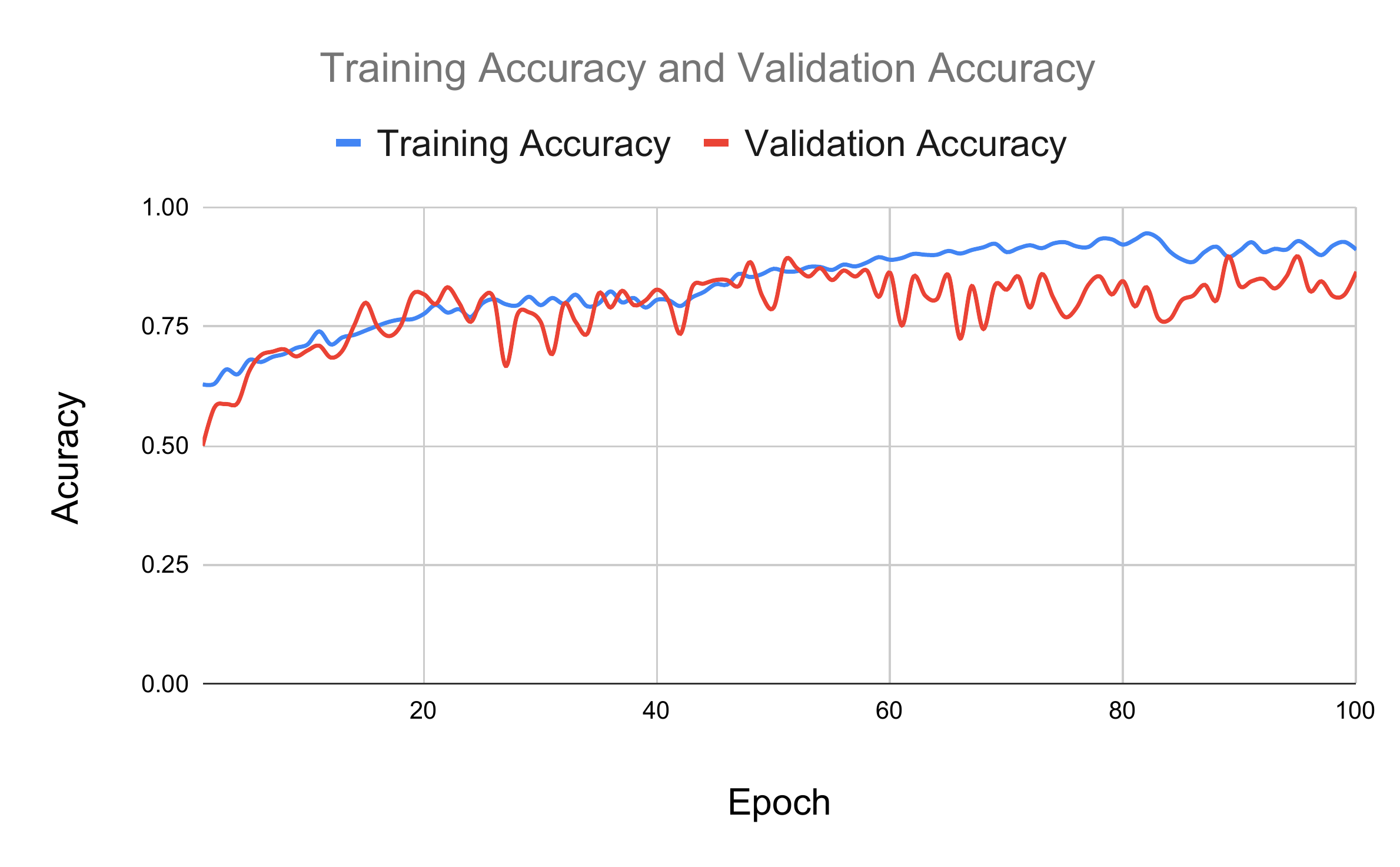}
        \label{fig:accuracy}
    }
    \\
    \subfigure[Comparison between training and validation loss]
    {
        \includegraphics[width=0.9\linewidth]{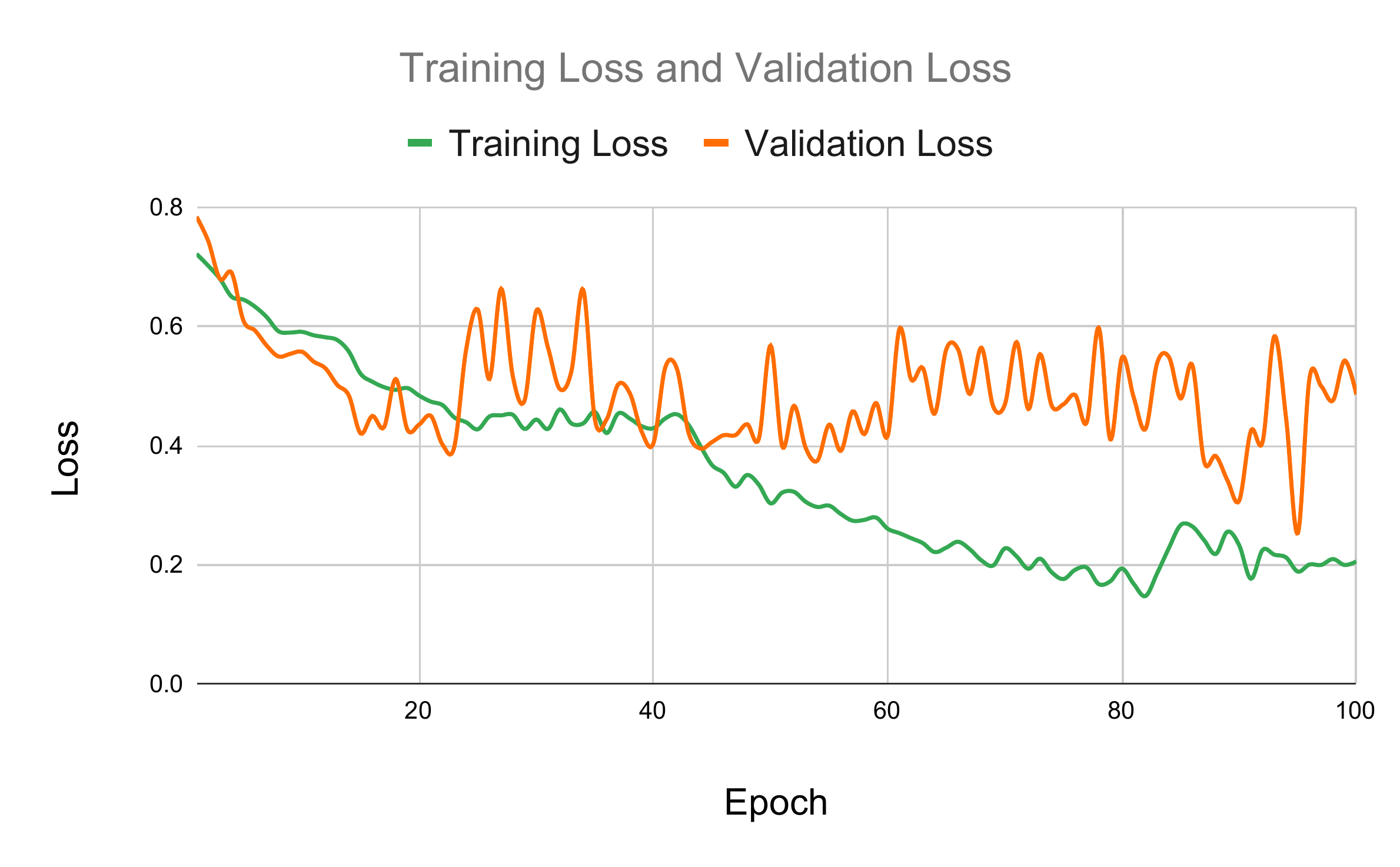}
        \label{fig:loss}
    }
    \caption{Performance measurement between training and validation process.}
    \label{fig:mdcn_train_vs_val}
\end{figure}

	\subsection{Qualitative Analysis}
We illustrate visualization of features from an individual DSTC block and compare with the features from a single 3D CNN layer, which is used as a branch in our model. 
In the Fig.~\ref{fig:features_vis}, we show three consecutive frames of samples from RWF-2000 dataset denoted by $F_{t-1}$, $F_t$, and $F_{t+1}$. 
We also illustrate the features extracted from the first DSTC block and compare them with the single 3D CNN layer. 
From the visualization, it is evident that the combination of 1D, 2D, and 3D CNN layers extract salient features from the input and makes our model efficient and accurate.

    \subsection{Comparison with the state-of-the-art}
    
      In this section, we compare our models with other violence detection state-of-the-art models on the datasets stated earlier.
      
      Table~\ref{tab:hockey_movie} shows the comparison of our model with other models on Hockey-fight and Movie-fight dataset. 
      Our model outperforms the hand-crafted features based models and deep learning-based models as well on both datasets. 
      2s-MDCN obtained ${100.0\%}$ and ${99.0\%}$ accuracy on Hockey-fight dataset and Movies-fight dataset respectively.
      
      The comparison on the RWF-2000 violence dataset is shown in the table~\ref{tab:rwf-2000}. 
      Hence, we have also reported the number of parameters (M) and computational complexity along with  accuracy. 
      Computation complexity is expressed in GFLOPs ($ {10}^{9} $ FLOPs), where, 1 FLOP is defined as 1 floating-point multiple-addition operation~\cite{zhang2018shufflenet}.
      2s-MDCN (RGB only) achieved 87.50$\%$ accuracy only using RGB clips whereas, FlowGate (RGB) achieved 84.50$^\%$ with twice the complexity of our model. 
      Moreover, our model also outperformed FlowGate with the fusion of RGB frames and optical flow, though our model has a computational complexity of 4.47 GFLOps whereas FlowGate performs 16.98 GFLOPs. 
      2s-MDCN (fusion) achieved an accuracy of 89.7\% with 0.92M parameters and computational complexity of 8.97 GFLOPs.
      2s-MDCN also outperformed other models listed in Table~\ref{tab:rwf-2000} in terms of accuracy, memory consumption and cost.
      Thus, our model achieves state-of-the-art accuracy with lower computational cost and less parameter size. 
      Furthermore, the exclusion of optical flow eliminates the overhead of pre-processing of input which made our model more efficient.
      
      Additionally, in Figure~\ref{fig:mdcn_train_vs_val}, a comparison between the loss and accuracy of the model in training and validation phase on RWF-2000 dataset is reported. 
    As illustrated in Figure~\ref{fig:accuracy}, accuracy was stable both in the training and validation process during the whole training. However, the loss shows a slight overfitting during epoch 25 to 35 as illustrated in Figure~\ref{fig:loss}. Later, it is fixed as training progress and the model learns the features.
      
      Unlike regular human action recognition, violence detection is more complicated because of the involvement of more motion, complex background, and dynamic of character varies more.
      In spite of being smaller in size, our model extracts information from this complex features and achieves state-of-the-art accuracy, and also can be used in real-world situations.
      

\begin{table}[htb]
\begin{center}
\begin{tabular}{lcc}
\hline
\multicolumn{1}{l}{\multirow{2}{*}{Model}} & \multicolumn{2}{l}{Processing Speed (FPS)} \\\cline{2-3}
\multicolumn{1}{c}{}                       & \multicolumn{1}{c}{CPU} & \multicolumn{1}{c}{Jetson Nano} \\\hline
R(2+1)D~\cite{r21dtran2018closer}                                    & 7.5                     & 13.0                   \\
C3D~\cite{c3dtran2015learning}                                        & 11.2                    & 18.8                   \\
I3D (two-stream)~\cite{i3dcarreira2017quo}                           & 3.68                    & 12.6                   \\
FlowGate (fusion)~\cite{cheng2019rwf}                          & 12.4            & 58.18                  \\ \hline
2s-MDCN (RGB only)                                & \textbf{16.6}                    & \textbf{80.0} \\
2s-MDCN (fusion)                                & \textbf{13.3}                    & \textbf{62.0} \\\hline                
\end{tabular}
\caption{Comaparison of processing speed on CPU and Jetson Nano.}
\label{tab:mdcn_fps}
\end{center}
\end{table} 

\subsection{Performance analysis of 2s-MDCN on edge devices}
    The performance of our model on edge devices is also shown to demonstrate the deployability of 2s-MDCN model in a real-time surveillance system.
    We show the processing speed in terms of frames per second (FPS), in Table ~\ref{tab:mdcn_fps}. We report FPS of our 2s-MDCN and other state-of-the-art models, both on a central processing unit (CPU) and an edge device called Jetson Nano.
    Our model can perform at 16.6 FPS on CPU and 80 FPS on Jetson Nano with only RGB frames, which makes our model more than 37$\%$ faster than FlowGate(fusion) on Jetson Nano. 
    When combined with optical flow, 2s-MDCN processes 13.3 frames per second on CPU and 62 frames on Jetson Nano, which makes it suitable for surveillance systems.
    Our 2s-MDCN model, therefore, performs better and faster with low computational cost, which makes it a viable choice for real-time violence detection applications.

    \section{Conclusion}
    In this work, we have represented a novel architecture for violence detection. 
    This model achieves state-of-the-art accuracy in the largest violence detection dataset. 
    This is possible by extracting efficient temporal and spatial features and reducing the loss of information during the prediction process.
    Our model outperformed FlowGate(fusion) with almost one-fourth of computational cost and better accuracy.
    As our model takes less memory, consumes less power, and removes pre-processing overhead, it becomes a suitable candidate for a violence detection model which can be deployed in real-life scenarios.

    
{\small
\bibliographystyle{ieee}
\bibliography{rmcn}
}

\end{document}